\pdfoutput=1

\documentclass[11pt]{article}

\usepackage[final]{acl}

\usepackage{times}
\usepackage{latexsym}

\usepackage[T1]{fontenc}

\usepackage[utf8]{inputenc}

\usepackage{microtype}

\usepackage{inconsolata}

\usepackage{graphicx}

\usepackage{todonotes}
\usepackage{linguex}
\usepackage{coptic}
\usepackage{multirow}
\usepackage{cjhebrew}
\usepackage{tikz-dependency}
\usepackage{caption}
\usepackage{subcaption}
\usepackage[export]{adjustbox}
\usepackage{epsfig}
\usepackage{stfloats}
\usepackage{booktabs}

\newcommand{\verteq}{\rotatebox{25}{=}}

\title{A UD Treebank for Bohairic Coptic}

\author{Amir Zeldes \\
  Georgetown University \\
  \texttt{amir.zeldes@georgetown.edu} \\\And
  Nina Speransky \\
  Hebrew University of Jerusalem \\
  \texttt{gitchinama@gmail.com} \\ \AND
  Nicholas Wagner \\
  Duke University \\
  \texttt{nicholas.wagner@duke.edu} \\\And
  Caroline T. Schroeder \\
  University of Oklahoma \\
  \texttt{ctschroeder@ou.edu} \\
  }

\begin{document}
\maketitle
\begin{abstract}
Despite recent advances in digital resources for other Coptic dialects, especially Sahidic, Bohairic Coptic, the main Coptic dialect for pre-Mamluk, late Byzantine Egypt, and the contemporary language of the Coptic Church, remains critically under-resourced. This paper presents and evaluates the first syntactically annotated corpus of Bohairic Coptic, sampling data from a range of works, including Biblical text, saints' lives and Christian ascetic writing. We also explore some of the main differences we observe compared to the existing UD treebank of Sahidic Coptic, the classical dialect of the language, and conduct joint and cross-dialect parsing experiments, revealing the unique nature of Bohairic as a related, but distinct variety from the more often studied Sahidic. 
\end{abstract}

\section{Introduction}

\subsection{Coptic}
Coptic was the indigenous spoken and written language of Egypt during the Late Roman, Byzantine, and early Islamic periods. As the final stage of the Ancient Egyptian branch of the Afro-Asiatic language family, Coptic concludes a linguistic tradition with the longest continuous written record in human history, which includes three millennia of Hieroglyphic,  Hieratic and Demotic Egyptian writing, as well as over a millennium of writing in Coptic itself, a form of the same language written mainly in Greek letters. 

\begin{figure}[b!t]
\centering
\includegraphics[width=0.35\textwidth,frame]{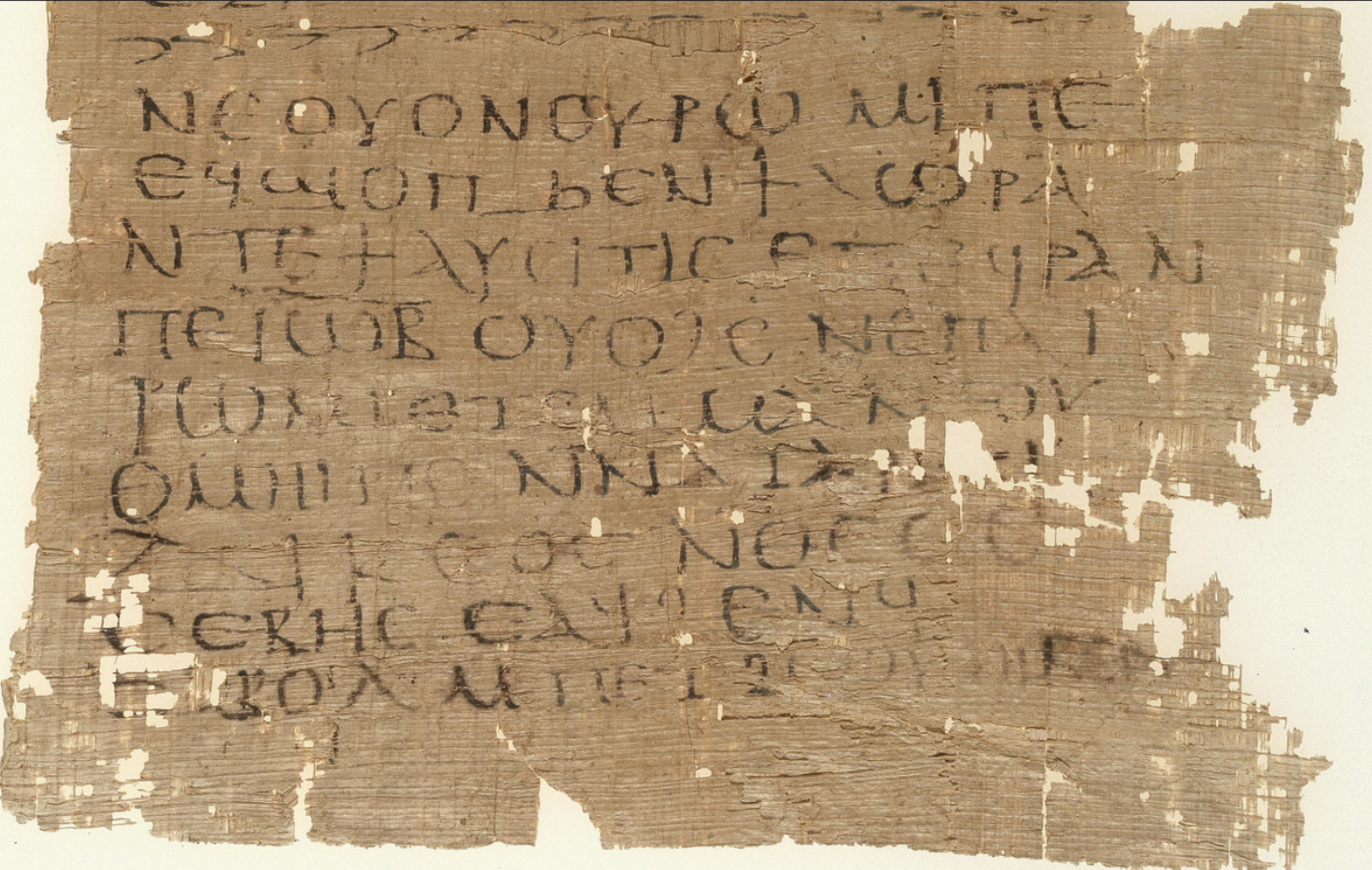}
\vspace{-6pt}
\caption{Excerpt from a papyrus containing Bohairic Job 1:1 (P.Mich.inv. 926 recto, TM no. 107875). Image: University of Michigan Library Digital Collections.}
\label{fig:ms}
\vspace{-14pt}
\end{figure}

Initially a very low resource language, recent efforts to digitize and annotate data for Coptic have resulted in now substantial resources for Sahidic Coptic, the classical dialect of the language, with corpora \cite{SchroederZeldes2020}, 
a machine readable dictionary \cite{FederKupreyevManningEtAl2018} and a UD dependency treebank \cite{zeldes-abrams-2018-coptic}. At the same time, other forms of Coptic, namely dialects beyond Sahidic, continue to have little or no annotated resources. In this paper, we aim to address this gap by introducing a UD treebank for a second, very significant dialect of the language: Bohairic Coptic. In this section we offer a brief summary of Coptic and its dialects, highlighting especially some of the main differences between the classical Sahidic, and Bohairic Coptic, which we explore in more detail later on using our data.


Geographic diversity in late ancient Egypt resulted in a range of regional dialects rather than one standardized form of Coptic. Scholars identify six principal dialects, but two of these, Sahidic and Bohairic, were the most influential \cite{Kasser1991}. Sahidic dominated as the literary language from the third to the ninth century CE before being gradually replaced by Bohairic, a northern variety that continues to serve today as a heritage and liturgical language among Coptic communities in Egypt and the diaspora. Despite the diversity of surviving evidence for Coptic dialects, print and digital resources for the language have remained limited and have focused almost exclusively on Sahidic. 
The textual evidence for the dialect of Bohairic, though sizeable (and primarily literary), is accessible mostly in facsimiles (see Figure \ref{fig:ms} for a papyrus manuscript example) or in older print publications, and many works remain unpublished \cite{ShishaHalevy2007}. Using terms from \citet{joshi-etal-2020-state}, Bohairic is at rank 0 of the language technology hierarchy, a `left-behind' language.

\subsection{Bohairic and Sahidic}
Expanding digital corpora to include Bohairic texts presents substantial challenges, as existing tools for even basic preprocessing operations, such as word segmentation and lemmatization, let alone syntactic parsing, require tools trained on dialect-specific data (see Section \ref{sec:parsing} for some evaluation). Although Sahidic and Bohairic are dialectal manifestations of a single language -- sharing a broadly consistent grammatical architecture and much of their lexical inventory, their phonological, orthographic and morphosyntactic systems diverge in ways that make NLP tools trained on Sahidic unsuitable for processing Bohairic texts.

At the most elementary level, the two dialects diverge already in their orthographic systems. Unlike earlier Egyptian -- written in hieroglyphic and hieratic scripts over the preceding three millennia -- Coptic adopted a modified version of the Greek alphabet, with additional characters of Demotic origin (ultimately derived from hieroglyphs) to represent sounds absent from Greek. Both dialects make use of six such additional letters, including the letter hore (\begin{coptic}h1\end{coptic}, Unicode U+03E9 lowercase / U+03E8 uppercase) for the voiceless glottal fricative /h/, but Bohairic distinguishes hore /h/ from khei /x/ (\begin{coptic}j\end{coptic} , Unicode U+2CC9 lowercase / U+2CC8 uppercase). This distinction is semantically consequential --  while the Sahidic word \begin{coptic}eh1rai\end{coptic} (ehrai) can confusingly mean both `up' or `down', and can only be disambiguated in context, in Bohairic we see distinct forms that had merged in Sahidic: \begin{coptic}ejrhi\end{coptic} (exr\={e}i) `down' and \begin{coptic}eh1rhi\end{coptic} (ehr\={e}i) `up'.

As an agglutinative language, Coptic combines multiple morphosyntactic elements into units known as bound groups. Following modern editorial conventions, these groups are defined by the presence of a single stressed lexical item at their core \cite{Layton2011}. Material that would normally be tokenized into separate words in annotated corpora often appears conflated into a single space-delimited string in Coptic. Such fusion is common, for example, in noun phrases or prepositional phrases, as in \ref{ex:np-pp}, or in auxiliaries and clitics attaching to verbs, as in \ref{ex:aux-clitic}, both examples in the Bohairic dialect.

\exg.\begin{coptic}h1iten\0p1\0ouwy\end{coptic}{\quad}\begin{coptic}m\0p1-noutj\end{coptic} \\
hiten-p$^{\mathrm h}$-w\={o}\v{s}{\quad}{\quad~~}m-p$^{\mathrm h}$-nuti{\quad} \\
by-the-will{\quad}{\quad~~}of-the-god \\
``by the will of God'' \label{ex:np-pp}

\exg.\begin{coptic}a\0c\08re\0f\0nah1m\0ou\end{coptic} \\
a-s-t$^{\mathrm h}$re-f-nahm-u \\
\textsc{past-3.sg.f-caus-3.sg.m}-hear-\textsc{3.pl} \\
``She made him save them'' \label{ex:aux-clitic}

In this paper we use hyphens to split such space-delimited word forms into tokens, which correspond to units that can receive independent parts of speech, such as nouns and verbs, articles and prepositions, etc. Note that such separators do not exist in source texts and are represented in the treebank directly through the tokenization. For words containing derivational affixes as in \ref{ex:morph}, and compound nouns or verbs \ref{ex:compound}, we additionally provide a segmentation into component parts in an additional annotation called \textsc{MSeg} in the conllu format MISC field, following existing practices in UD treebanks.\footnote{See \url{https://universaldependencies.org/misc.html\#mseg}}

\vspace{-6pt}
\exg. \begin{coptic}metatdjwnt\end{coptic} \\
MSeg=\begin{coptic}met\0at\0djwnt\end{coptic} \\
met-at-\v{c}\={o}nt \\
less-ness-anger \\
``angerlessness''\label{ex:morph}

\exg. \begin{coptic}erh1al\end{coptic} \\
MSeg=\begin{coptic}er\0h1al\end{coptic} \\
er-hal \\
do-service \\
``serve''\label{ex:compound}

Such words are easy to recognize since they have distinct, recurring forms (known affixes, special reduced forms of verbs with incorporated objects), lack internal syntactic markers (e.g.~incorporated objects like \textit{hal} in \ref{ex:compound} appear without articles or other modifiers, the verb `ire'  meaning ``do'' is reduced to `er') and are considered single nouns or verbs in terms of parts of speech, as well as single dictionary entries in terms of lemmatization. Due to these multiple levels of complexity, bound groups must first be analyzed and segmented before we can digitize Coptic texts in a way that allows for searchability. Each token can then be lemmatized and tagged to enable structured querying and lexical lookup, or linking to resources such as the Coptic Dictionary Online (CDO, \citealt{FederKupreyevManningEtAl2018}). 

\section{Previous work}

The Bohairic UD treebank joins a growing body of typologically diverse languages analyzed using the UD framework \cite{MarneffeEtAl2021}, including recent treebanks of related Afro-Asiatic languages such as Biblical and Modern Hebrew \cite{swanson-tyers-2022-universal,zeldes-etal-2022-second}, Arabic \cite{taji-etal-2017-universal}, and very recently, Ancient Egyptian \cite{antonio-diaz-hernandez-carlo-passarotti-2024-developing}, resources we consider in the development of comparable annotation guidelines (and to a lesser extent, treebanks converted to UD, e.g.~for Hausa, \citealt{Caron2015}).

The most important previous resource we model our work on is the existing UD treebank for Sahidic Coptic \cite{zeldes-abrams-2018-coptic}, which contains around 60K words from a range of works in a number of genres. In particular, the Sahidic treebank contains some Biblical material which is in part also available in Bohairic (see Section \ref{sec:data} below). By selecting the same Biblical books and chapters where possible, we are able to conduct some direct comparisons between the dialects which only target parallel passages (see Section \ref{sec:comparing}). At the same time, for texts in the Sahidic treebank that are unavailable in Bohairic, we select substitutes from similar genres, offering a similar range of language usage.

In terms of annotation scheme, we closely follow and adapt to Bohairic the Coptic Scriptorium guidelines for annotating parts of speech, lemmatization, sentence splitting and UD dependency relations. In a recent paper, \citet{Crellin2025} criticizes the choice of UD as a treebanking framework, among other languages for Coptic, as unmotivated, stating ``no overt discussion of the rationale for choosing [...] Universal Dependencies'' could be found \cite[100]{Crellin2025}. We would therefore like to explicitly motivate the choice of UD for the Bohairic Treebank beyond the obvious benefit of comparability with other resources (see Section \ref{sec:comparing}), and outline some of the key decisions in treebanking Coptic.

The most crucial decision we follow is treatment of lexical verbs as heads of their clauses, despite their etymology in many tenses in earlier Egyptian as subordinated, nominalized infinitives. For example, the verb ``hear'' in \ref{ex:durative} is the only possible head for the clause, which otherwise contains only the subject. Meanwhile in \ref{ex:tripartite}, the past auxiliary `a' in `a-f-s\={o}tem' would have been analyzed as the head in earlier Late Egyptian, as the construction is derived from a periphrasis, Late Egyptian \textit{jr\verteq{f} s\underline{d}m}, lit. ``did-he hear'', or more freely, ``he did hearing''.

\vspace{-6pt}
\exg. \begin{coptic}f\0cwtem\end{coptic} \\
f-s\={o}tem \\
he-hear \\
``he hears''\label{ex:durative}

\exg. \begin{coptic}a\0f\0cwtem\end{coptic} \\
a-f-s\={o}tem \\
\textsc{pst}-he-hear \\
``he heard''\label{ex:tripartite}

We motivate the choice of the lexical verb as the head by the basic UD principle of lexico-centrism, which also provides a more parallel analysis for subjects in the durative and past tenses shown above. Since Coptic, synchronically an agglutinative language, has a broad range of tenses and constructions, but only a handful of etymological sources for agglutinative morphemes (almost always forms of the Late Egyptian verbs for ``do'', ``give'' or ``know''), choosing a non-lexico-centric scheme would result in an analysis in which Coptic has only a handful of distinct verbs.
The choice of UD is therefore quite conscious, and especially given the benefits of comparability, without obvious superior alternatives.

We also match our native Coptic part of speech tagging scheme and its mapping to Universal POS tags to those used in the Sahidic treebank, as well as using Multiword Tokens (MWTs) to represent bound groups in the conllu format (i.e.~examples \ref{ex:durative} and \ref{ex:tripartite} would be one MWT each, containing two and three word forms respectively).

\section{Data}\label{sec:data}

The textual data in the Bohairic UD corpus consists of selections from works in multiple genres. We include selections from hagiography (saints' lives) of two prominent figures in early Christian Egypt: Shenoute, the leader of a federation of male and female monasteries in the fourth-fifth centuries; and Isaac, Patriarch (or Coptic Pope) of Alexandria from 686 to 689 CE. The Life of Shenoute is a compilation of panegyric works about Shenoute written in Bohairic Coptic over decades after his death and compiled into the genre of a saint's life (see \citealt{Lubomierski2008} and \citealt{Berno2019shenoute}). The Life of Isaac is a saint's life written in Bohairic Coptic possibly in the seventh century after Isaac's death, focusing mostly on his adult life as a monk, priest, and patriarch during the early Islamic period \cite{Berno2019isaac}. The Lausiac History is a narrative text originally written in Greek consisting of anecdotes about fourth-century monks and so has hagiographical elements as well as generic elements from travel narratives \cite{Berno2019lausiac}. Three biblical texts are also translations from Greek: the Gospel of Mark and 1 Corinthians from the New Testament, and the Christian Old Testament book of Habakkuk. Bohairic Habakkuk is likely a translation from the Septuagint (the Greek version of the Hebrew Bible). Mark is a gospel or ancient biography, and 1 Corinthians is a letter by Paul the apostle, meaning our Biblical data also spans multiple genres.

The text of each of these works is derived from previous digital or print editions. The \textit{Life of Shenoute} comes from a digital edition created by Hany Takla of the St. Shenouda the Archimandrite Coptic Society, based primarily on the edition published by \citet{Leipoldt1906}. Dr. Lydia Bremer-McCollum of the Coptic Scriptorium project\footnote{\url{https://copticscriptorium.org/}} extracted digitized text using optical character recognition (OCR) from the public domain edition of the  \textit{Life of Isaac} and the public domain edition of the Bohairic \textit{Lausiac History}, both edited by \citet{Amelineau1887,Amelineau1890}, followed by manual correction of OCR errors. The Gospel of Mark and 1 Corinthians digital texts come from the Marcion Project;\footnote{\url{https://marcion.sourceforge.net/}} both ultimately derive from the print edition of \citet{Horner1905}. The text of Habakkuk is from the working digital edition in-progress created by the G\"{o}ttingen Coptic Old Testament project.\footnote{\url{https://coptot.manuscriptroom.com/}} All of the digital versions were processed by the Coptic Scriptorium project's natural language processing tools \cite{zeldes-schroeder-2016-nlp}, which have normalized the text, including normalizing variant spellings and expanding any abbreviations; for this paper, all NLP processing has been manually checked by one or more of the authors before treebanking.

\begin{table*}[t!bh]
\centering
\begin{tabular}{l|lrrr}
\toprule
\textbf{source}  & \textbf{genre}     & \textbf{chapters} & \textbf{tokens} & \textbf{sentences} \\
\midrule
1 Corinthians    & Biblical Epistle   & 1--7               & 4,789            & 164                \\
Gospel of Mark   & Biblical Narrative & 1--9               & 11,091           & 373                \\
Book of Habbakuk & Poetic             & 1--3               & 1,988            & 56                 \\
Life of Shenoute & Hagiography        & 1--26              & 4,970            & 148                \\
Life of Isaac    & Hagiography        & 1--19              & 5,433            & 143                \\
Lausiac History  & Hagiography        & 1--16              & 4,453            & 117               \\
\midrule
\textbf{Total:} & & & 32,724 & 1,001 \\
\bottomrule
\end{tabular}
\caption{Data in the Bohahiric Treebank.}\label{tab:corpus}
\end{table*}

\subsection{Inter-annotator agreement}
To evaluate the quality of our annotations, we double-annotated 166 sentences (6,207 tokens) from two different texts, the Life of Shenoute and the Lausiac History. Table \ref{tab:agreement} shows the results for Cohen's $\kappa$ and mutual F1 score, for both dependency relations (labels) and dependency heads. In order to avoid inflating $\kappa$ due to a large range of possible labels for heads (i.e.~all numbers attested as dependency heads), we represented heads as offsets from the position of the child (i.e.~if token 37 has token 35 as its parent, we tally the value as -2). This increases the probability of chance agreement and prevents an inflated metric due to label proliferation. For dependency relations, we use the full label including subtypes (see Appendix \ref{sec:appendix-labels} for details).

\begin{table}[h!tb]
\resizebox{\columnwidth}{!}{%
\begin{tabular}{l|rrrr}
\toprule
\textit{}                 & \multicolumn{2}{c}{\textbf{Labels}}                                         & \multicolumn{2}{c}{\textbf{Heads}}                                          \\
                          & \multicolumn{1}{l}{\textbf{Kappa}} & \multicolumn{1}{l}{\textbf{F1}} & \multicolumn{1}{l}{\textbf{Kappa}} & \multicolumn{1}{l}{\textbf{F1}} \\
                          \midrule
\textit{Life of Shenoute} & 92.91                              & 93.49                                  & 93.84                              & 94.77                                  \\
\textit{Lausiac History}  & 95.53                              & 95.78                                  & 94.61                              & 95.53                                  \\
\midrule
\textbf{Macro average}    & 94.22                              & 94.64                                 & 94.23                             & 95.15                                  \\
\textbf{Micro average}    & 94.79                              & 95.12                                  & 94.39                              & 95.29    \\
\bottomrule

\end{tabular}
}
\caption{Cohen's Kappa ($\kappa$) and mutual F1 score in two texts (166 sentences) for dependency relation labels and heads, the latter represented as offsets relative to child tokens.}
\label{tab:agreement}
\end{table}

The results show very high agreement, substantially in excess of initial (pre-adjudication) scores in the 80s for the original version of the Sahidic treebank \cite[199]{zeldes-abrams-2018-coptic}. This is likely due to the fact that annotators in this case were post-graduate researchers with substantial Coptic annotation experience, as opposed to the novice student scores reported in the Sahidic paper.

The data above constitutes the first openly available morphosyntactically annotated corpus of Bohairic, and allows for a number of quantitative comparisons with the Sahidic, to which we now turn below.

\section{Comparing UD Bohairic and Sahidic}\label{sec:comparing}

Like other projects adding UD treebanks in low-resource languages for which closely related languages already have treebanks \cite{jobanputra-etal-2024-universal}, one of our goals is to explore the ways in which Bohairic diverges from other varieties of Coptic -- ideally, we would like to have treebanks of all Coptic dialects, but for the present we must focus on the comparison with Sahidic. 

Dialects and closely related languages can differ in two different ways: they can have categorically distinct constructions, such as different auxiliaries, distinct argument structures for equivalent verbs etc., or they can use similar constructions but in quantitatively different usage patterns. While categorical differences between Bohairic Coptic and the better studied Sahidic are relatively well understood, quantitative differences are more elusive, but can show up in a corpus analysis.

On the lexical level, we can note that of the approximately 2,800 unique words attested in the Sahidic treebank, and around 2100 unique words in the Bohairic treebank, only about 600 are shared, and even among these, identical forms do not always translate to identical meanings. For example while \begin{coptic}et\0\end{coptic} `et-' can be a form of the relative marker in both Sahidic and Bohairic, it can also represent the precursive form meaning `after' in Bohairic. Among the disjoint, dialect-specific vocabulary, many words have corresponding words in the other dialect which differ only due to pronunciation, but some words are totally unique, such as \begin{coptic}baki\end{coptic} `baki', which means `town' in Bohairic, but does not exist in Sahidic.

On the syntactic level, we cannot find differences and commonalities as easily, but thanks to the existence of UD trees in both dialects, we can still leverage annotated data in a straightforward way. To find some of the clearest differences that our data reveals, we extract relative proportions of the dependency relations attested in the Bohairic and Sahidic Coptic treebanks, an excerpt of which is presented in Table \ref{tab:ratios}, which is sorted by the ratio of frequencies.

\begin{table}[h!tb]
\resizebox{\columnwidth}{!}{%
\begin{tabular}{l|rr|rr|r}
\toprule
                    & \multicolumn{2}{c|}{\textbf{Sahidic}}                                     & \multicolumn{2}{c|}{\textbf{Bohairic}}                                    & \multicolumn{1}{l}{}               \\
\textit{}           & \multicolumn{1}{l}{\textbf{count}} & \multicolumn{1}{l|}{\textbf{per 1K}} & \multicolumn{1}{l}{\textbf{count}} & \multicolumn{1}{l|}{\textbf{per 1K}} & \multicolumn{1}{l}{\textbf{ratio}} \\
\midrule
\textit{iobj}       & 84                                 & 1.471                            & 36                                 & 1.100                            & 0.747                           \\
\textit{…}          &                                    &                                     &                                    &                                     &                                    \\
\textit{cop}        & 500                                & 8.757                            & 291                                & 8.892                            & 1.015                           \\
\textit{nsubj}      & 5,549                               & 97.185                            & 3,275                               & 100.073                            & 1.029                           \\
\textit{…}          &                                    &                                     &                                    &                                     &                                    \\
\textit{dislocated} & 889                                & 15.569                            & 670                                & 20.473                            & 1.314\\
\bottomrule
\end{tabular}
}
\caption{Frequencies and their ratios for some Bohairic and Sahidic dependency relations.}
\label{tab:ratios}
\end{table}

As the table shows, striking differences are present for example in the frequencies of \texttt{dislocated} arguments (much more common in Bohairic) and \texttt{iobj} (indirect objects, much more common in Sahidic). We also note that some control labels, which we would expect to align across datasets, are quite comparable, such as \texttt{cop} for copulas, or \texttt{nsubj} for subjects.

\subsection{Subject dislocation}

While all Coptic dialects use a basic SVO word order for tensed clauses with lexical verbs \cite{Loprieno2000}, both left and right dislocations are well attested, with left-dislocation of any argument (e.g.~subject or object) bearing no special marking, as in \ref{ex:disloc}.\footnote{In the following, some examples from the treebank are abbreviated for space and clarity.} Here again, the choice of UD (contra \citealt{Crellin2025}) means we can easily find these, thanks to the UD label \texttt{dislocated}.

\exg.\begin{coptic}pi\0sjai\end{coptic}{\quad}{\quad}{\quad}\begin{coptic}gar\end{coptic}{\quad}\begin{coptic}f\0jwteb\end{coptic} \\
pi-sxai{\quad}{\quad}{\quad}gar{\quad}~~f-x\={o}teb \\
the-scripture for{\quad}~it-kill \\
``For scripture, it kills'' (i.e.~scripture kills)\label{ex:disloc}

By contrast, right dislocation or extraposition is obligatorily marked for subjects using the particle \begin{coptic}nd1e\end{coptic} `n\v{c}e' in Bohairic, paralleled in Sahidic by the form \begin{coptic}nqi\end{coptic} `nk$^{\mathrm j}$i'. This particle has been analyzed as a post-verbal nominative case marker by \citet{Grossman2015}, who notes that ``postverbal subjects are more frequently new referents'' in Sahidic, but rarely so in Bohairic. Examples \ref{ex:nqi}--\ref{ex:nje} demonstrate the use of the marker for right dislocation in Sahidic and Bohairic respectively:

\exg. \small \begin{coptic}a\0f\0bwk\end{coptic}{\quad}\begin{coptic}eh1oun\end{coptic}{\quad}\begin{coptic}e\0pef\0hi\end{coptic}{\quad}~~~\begin{coptic}nqi\0iwh1annhc\end{coptic}{\quad}\\
a-f-b\={o}k{\quad}ehoun{\quad}e-pef-\={e}i{\quad}~~~nk$^{\mathrm j}$i-i\={o}hann\={e}s\\
\small PST-he-go{~~}inside{\quad}to-his-house{\quad}PTC-John \\
``He went into his house, (that is) John''\label{ex:nqi}

\exg. \small \begin{coptic}a\0f\0qici\end{coptic} \begin{coptic}nd1e\0p1\0rh\end{coptic} \\
a-f-k$^{\mathrm j}$isi n\v{c}e-ph-r\={e} \\
\small PST-it-exalt PTC-the-sun \\
``It was exalted, (that is) the sun''\label{ex:nje}

One question we can immediately explore using our data and the existing UD Sahidic treebank is whether the constructions are used comparably often in the two dialects. Since segmentation, tagging and parsing guidelines match across the treebanks, we can be confident that all relevant cases can be found using equivalent searches -- the results are shown in Table \ref{tab:nqi}.

\begin{table}[h!tb]
\resizebox{\columnwidth}{!}{%
\begin{tabular}{l|rrr}
\toprule
                  & \textbf{matches} & \textbf{total words} & \textbf{freq per 1K} \\
                  \midrule
& \multicolumn{3}{c}{\textbf{All data}} \\
\midrule
\textit{Boh.} \begin{coptic}nd1e\end{coptic} & 258              & 32,724                & 7.88              \\
\textit{Sah.} \begin{coptic}nqi\end{coptic} & 206              & 57,097                & 3.61  \\
                  \midrule
& \multicolumn{3}{c}{\textbf{Parallel only}} \\
\midrule
\textit{Boh.} \begin{coptic}nd1e\end{coptic} & 122              & 15,880                & 7.68              \\
\textit{Sah.} \begin{coptic}nqi\end{coptic} & 66              & 15,619                & 4.22  \\
\bottomrule
\end{tabular}
}
\caption{Bohairic and Sahidic frequencies for right dislocated subjects.}
\label{tab:nqi}
\vspace{-6pt}
\end{table}

The table shows that right subject dislocation is much more common in Bohairic, with more hits in total in the smaller Bohairic treebank, and more than double the relative frequency. The bottom of the table shows results for the same queries, restricted to only the parallel data available in both corpora -- these numbers are slightly less reliable since they are based on less material, but should be expected to be much closer since they derive from the same chapters of the Bible. Here too, the gap remains very substantial, despite the fact that if anything Biblical style should represent a conservative and less free or colloquial style.\footnote{An anonymous reviewer has also asked about comparable breakdowns by genre, but we feel that for some of the texts, comparisons would be hard to establish, and these would also split the already small treebank to the point where counts would be substantially less reliable.}

Looking at the Sahidic data in more detail, it becomes clear that the construction responsible for the discrepancy is nominal subjects in the canonical position between auxiliaries and verbs, as in \ref{ex:subj-canon-bible}, which is paralleled in Bohairic by \ref{ex:subj-disloc-bible} -- both examples render 1 Corinthians 2:10.

\exg. \small \begin{coptic}a\0p\0noute\end{coptic}{\quad}\begin{coptic}~qolp\0ou\end{coptic}{\quad}~~\begin{coptic}na\0n\end{coptic}{\quad}~~\begin{coptic}ebol\end{coptic}{\quad} \\
a-p-noute{\quad}~~k$^{\mathrm j}$olp-u{\quad}~~~na-n{\quad}~~ebol\\
\small PST-the-god{\quad}reveal-3Pl{\quad}to-1Pl{\quad~}out \\
``God revealed them to us''\label{ex:subj-canon-bible}.

\exg. \small \begin{coptic}a\0f\0qorp\0ou\end{coptic}{\quad\quad}\begin{coptic}na\0n\end{coptic}{\quad}\begin{coptic}ebol\end{coptic}{\quad}\begin{coptic}nd1e\0p1\0noutj\end{coptic}{\quad} \\
a-f-k$^{\mathrm j}$orp-u{\quad\quad}na-n{\quad}ebol{\quad}n\v{c}e-p$^{\mathrm h}$-nuti\\
 \tiny{PST-3Sg-reveal-3Pl}\small{\quad\quad}to-1Pl{\quad}out{\quad}PTC-the-god \\
``He revealed them to us, that is God''\label{ex:subj-disloc-bible}.

This construction is much rarer in Bohairic, and indicates that the Bohairic data represents a further step in the grammaticalization of the pronominal subject + auxiliary group, which forces subjects to be realized before or after the verbal complex. This tendency is well known from other languages in the Afro-Asiatic family, such as Hausa, which has similar subject + auxiliary structures, but can only realize a nominal subject outside of the verbal complex, with a `duplicate' pronoun mirroring the subject -- a pronominal TAM marker within the verbal complex is mandatory (see \citealt[331]{crysmann2012hag}, and \citealt{hartmann2006focus} on fronting in Hausa).

\subsection{Indirect objects}

Following the Sahidic treebank, we use the \texttt{iobj} label primarily to indicate the possessor in the predicative possession constructions with the predicates \begin{coptic}ouon\end{coptic}(\begin{coptic}te\end{coptic}) ``there exists'' and \begin{coptic}mmon\end{coptic}(\begin{coptic}te\end{coptic}) ``does not exist'', which are used with a subject expressive the possessum. Thus the Sahidic construction in \ref{ex:sah-ounte} with the corresponding Sahidic form \begin{coptic}ounte\end{coptic}  represents an oblique predication ``exists to your father something except sins'' (or etymologically more precisely, ``to the hands of your father'').

\exg. \small \begin{coptic}ounte\0pek\0eiwt\end{coptic}{\quad}\begin{coptic}laau\end{coptic}{\quad}\begin{coptic}nsa\0h1en\0nobe\end{coptic} \\
wnte-pek-i\={o}t{\quad}{\quad}la'u{\quad~~}nsa-hen-nobe \\
\small EXIST-your-father ~thing{\quad}beyond-some-sin \\
``Does your father have anything except sins?''\label{ex:sah-ounte}

Although Sahidic also prefers pronouns to nouns in the possessor position, this is more extremely the case in Bohairic. What is more, although the same constructions as in Sahidic are possible in Bohairic, the Bohairic data shows a tendency to postpone possessors to a later prepositional phrase, leaving the indirect object slot immediately after the existence predication empty. For example, we can contrast the constructions from Mark 4:9 in Bohairic \ref{ex:boh-ounte-mark} and Sahidic \ref{ex:sah-ounte-mark}:

\exg. \small \begin{coptic}p1h\end{coptic}{\quad~~}\begin{coptic}ete\0ouon\end{coptic}{\quad}\begin{coptic}ou\0mayd1\end{coptic}{\quad}\begin{coptic}mmo\0f\end{coptic}{\quad} \\
ph\={e}{\quad}ete-won{\quad~~}u-ma\v{s}\v{c}{\quad\quad~}mmo-f\\
\small \textsc{dem}{~~~}\textsc{rel-exist}{~~~}ear{\quad\quad\quad\quad}of-him \\
``he who has ears (to listen, let him listen)''\label{ex:boh-ounte-mark}.

\exg. \small \begin{coptic}p\0ete\0unt\0f\end{coptic}{\quad\quad\quad}\begin{coptic}maadje\end{coptic}{\quad}\begin{coptic}mmau\end{coptic} \\
p-ete-wnt-f{\quad\quad\quad~}ma'\v{c}e{\quad~~}mmau\\
\small the-\textsc{rel-exist}-him{~~}ear{\quad\quad\quad~}there \\
``he who has ears (to listen, let him listen)''\label{ex:sah-ounte-mark}.

The postponed cases of possessors mediated by prepositions thus appear to be the main driver of the lower frequency of \texttt{iobj} dependencies in the Bohairic data. As far as we are aware, this finding has not been published on to date.

\subsection{Focus and preterit marking}

Coptic belongs to the group of languages that employ morphological devices known as `converters' \cite[319-366]{Layton2011}, which are applied to entire clauses, converting them for example into information structurally marked focalized clauses (focus conversion), signal anterior past tense (the preterit conversion, creating imperfect readings from present clauses or pluperfects from perfect clauses), among others. 

For focus marking, two competing strategies are found: Cleft Sentences as in \ref{ex:cleft} and the morphological focus converter (sometimes referred to as the Second Tense, where the focalized present tense is called the `Second Present' etc.) as in \ref{ex:cfoc}.

\exg. \small \begin{coptic}ou\end{coptic}{~}\begin{coptic}p\0et\0ou\0iri\end{coptic}{\quad~}\begin{coptic}mmo\0f\end{coptic}{~}\begin{coptic}jen-ni-cabbaton\end{coptic} \\
ou{~~}p-et-ou-iri{\quad~~~}mmo-f{~~}xen-ni-sabbaton\\
\small{what}{~}\textsc{cop-rel-3pl}-do{~~}\textsc{acc}.it{~~}on-the.PL-sabbath\\
``What is it that they do on Sabbaths?''\label{ex:cleft}

\exg. \small\begin{coptic}et\0a\0i\0tj\0wmc\end{coptic}{\quad\quad~~}\begin{coptic}nw\0ten\end{coptic}{~~}\begin{coptic}jen-ou-mwou\end{coptic} \\
et-a-i-ti-\={o}ms{\quad\quad~~~~}n\={o}-ten{\quad~}xen-u-m\={o}w\\
\scriptsize \textsc{foc-pst-1sg}-give-baptism{\quad}to-you{\quad\quad~}in-a-water\\
\small``I christened you with WATER''\label{ex:cfoc}

The presence of the morpheme glossed as \textsc{foc} in \ref{ex:cfoc} indicates that a constituent is focalized in the sentence, in this case `water' (`I christened you with WATER' rather than something else). 

It has been observed that these strategies are represented unequally in Sahidic and Bohairic, with the preference for the focalizer in the former dialect, and for the cleft sentence in the latter \cite{Mueller2021}. However, up until now such observations have not been backed up with precise and reproducible quantitative data. The UD treebanks make it possible to find the Sahidic-to-Bohairic ratio of focus markers in various parts of the New Testament texts. Since the use of these markers is heavily context and content dependent, we restrict our search to just the Biblical sources available in both treebanks: For the Gospel of Mark, this ratio roughly equals 2.44, and for 1 Corinthians, it is 3.58, both clearly favoring the prevalence of the focalizer in Sahidic.

A less stark, but more surprising result can be found for the preterit marker in the two dialects: as the only formal category denoting anterior past, we could logically expect its equal representation in identical texts in Sahidic and Bohairic. Yet, the UD treebank statistics show that the preterit marker occurs in Bohairic almost twice as often as in Sahidic (the ratios for Mark and 1 Corinthians are 2.21 and 1.76 favoring Bohairic, respectively). These numbers show that there is a substantial difference in how the tense systems of the two dialects are constructed, though we leave a more detailed study of what stands in place of the preterit in Sahidic to future work.

\section{Parsing}\label{sec:parsing}

\subsection{Cross-corpus experiments}

To evaluate the possibility of using the treebank to train an effective parser for Bohairic Coptic, we train and test several models in different scenarios:

\begin{enumerate}
    \item Training on just the Bohairic train-set
    \item Training on just the Sahidic train-set
    \item Joint training on the Bohairic and Sahidic UD treebank train-sets
    \item Balanced training on the smaller Bohairic train-set, and an equal amount of Sahidic data
\end{enumerate}

The balanced setting is meant to evaluate whether joint training is more feasible if we ensure Bohairic examples are not overwhelmed by the larger amount of data available for Sahidic. In this case we take care not to include the same document (e.g.~the same Bible chapter) from both dialects, and otherwise randomly select Sahidic documents from the appropriate partition, until the Bohairic data size has been reached.

To run the experiments, we use DiaParser \cite{attardi-etal-2021-biaffine}, a neural biaffine dependency parser, using the default hyperparameters (see the Appendix for exact values). As input embeddings we use the MicroBERT architecture and Sahidic Coptic embeddings made available by \citet{gessler-zeldes-2022-microbert}, and train corresponding MicroBERT embeddings for Bohairic using all available Bohairic Coptic data from Coptic Scriptorium's online repository;\footnote{https://github.com/CopticScriptorium/corpora} the new Bohairic transformer embeddings will be released publicly via Hugging Face.

Table \ref{tab:parser-scores} shows the results for labeled and unlabeled attachment scores on the respective test sets in each setting, along with train and test partition sizes in thousands of tokens. The results initially reveal that, unsurprisingly, training and testing across dialects (red numbers) produces very poor results, with LAS and UAS scores around 73 and 62 respectively -- the scores are rather comparable in both directions, despite the availability of almost double the data when training on Sahidic. This indicates that the parser is unable to generalize when surface forms vary, since as we noted above, even auxiliaries and prepositions look quite different across dialects.

\begin{table}[t!b]
\resizebox{\columnwidth}{!}{%
\begin{tabular}{l|rrrr}
\toprule
\textbf{}         & \multicolumn{4}{c}{\textbf{test}}                                            \\
                  & \multicolumn{2}{c}{\textbf{Bohairic}} & \multicolumn{2}{c}{\textbf{Sahidic}} \\
                  & \multicolumn{2}{c}{\textbf{(11K tokens)}} & \multicolumn{2}{c}{\textbf{(10.3K tokens)}} \\
\textbf{train (tokens)}    & \textbf{LAS}      & \textbf{UAS}      & \textbf{LAS}     & \textbf{UAS}      \\
\midrule
\textit{Bohairic} (16.5K) & \textcolor{ForestGreen}{86.349}            & \textcolor{ForestGreen}{89.486}            & \textcolor{red}{62.205}           & \textcolor{red}{74.633}            \\
\textit{Sahidic} (35.8K)  & \textcolor{red}{62.683}            & \textcolor{red}{73.178}            & \textbf{\textcolor{ForestGreen}{89.760}}   & \textbf{\textcolor{ForestGreen}{92.489}}   \\
\textit{Joint} (52.3K)    & \textbf{89.929}   & \textbf{92.677}   & 88.449           & 91.602            \\
\textit{Balanced} (36K) & 88.989            & 91.927            & 86.858           & 90.628           \\
\bottomrule
\end{tabular}
}
\caption{Labeled (LAS) and Unlabeled (UAS) Attachment Scores in each setting when testing on each dialect. Within and across dialect scores are in \textcolor{ForestGreen}{green} and \textcolor{red}{red} respectively. The best settings for each dialect are bolded.}
\label{tab:parser-scores}
\end{table}

Single dialect models (green numbers) reveal a gap between the smaller Bohairic data and its larger Sahidic counterpart: while Sahidic obtains a LAS of 89.76, Bohairic lags behind with 86.349 (2.5 point difference), with an even starker difference in UAS (92.489 vs. 89.486, about 3 points). Given that the dialects and texts available in them are rather similar, this suggests that more Bohairic data is likely to have an impact in a single-dialect setting.

Moving to the joint models, both the balanced and full-joint scenario improve the score on Bohairic, suggesting that although word forms are different, syntactic structures are similar enough to generalize across the datasets. In fact, the \textsc{joint} setting outperforms \textsc{balanced}, suggesting that simply having more data is better, as long as there is a core of Bohairic examples to inform the parser about pivotal Bohairic word forms. In absolute terms, the joint Bohairic scores even slightly surpass the Sahidic single-dialect model scores, though these numbers are not strictly comparable, since the test sets contain different documents. We suspect this means the Bohairic test set may be slightly easier overall than the Sahidic one, but we also take it to be an indication that our annotations match the Sahidic guidelines closely.

By contrast to the Bohairic benefit from joint training, both joint scenarios perform worse on Sahidic than the Sahidic-only model. This suggests that given the amount of data available in Sahidic, the infusion of the smaller Bohairic data is not helpful. In the Sahidic case \textsc{balanced} is unsurprisingly the worse setting, since there is less total Sahidic data involved.

\subsection{Error analysis}

\begin{figure}[h!tb]
\centering
\includegraphics[width=0.5\textwidth, trim={21.5cm 0.4cm 10.5cm 13.2cm}, clip]{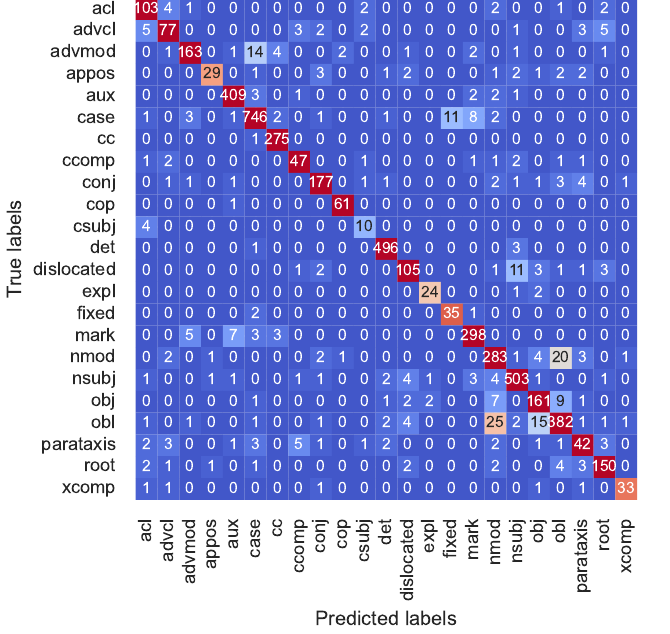}
\caption{Confusion matrix for collapsed major dependency relations for the Bohairic-only model on the Bohairic test set (labels with <10 occurrences are omitted).}
\vspace{-6pt}
\label{fig:confmat}
\end{figure}

To better understand what is challenging about Bohairic parsing, we perform quantitative and qualitative error analysis. Figure \ref{fig:confmat} shows the confusion matrix for dependency labels in the Bohairic test set for the Bohairic-only model (merging subtypes of the same major relations and omitting labels with fewer than 10 occurrences in the test set).

\vspace{-3pt}

As with most parsers, the most common confusion is between the \texttt{obl} and \texttt{nmod} labels, indicating problems with classic PP-attachment ambiguities. Overprediction of low adnominal attachment (\texttt{nmod} for true \texttt{obl}) is slightly more common than the opposite. Additional relatively common confusions occur for \texttt{dislocated} subjects versus regular subjects (\texttt{nsubj}), which is not surprising, and for \texttt{advmod} and \texttt{case} being confused with \texttt{case} and \texttt{fixed} respectively. The latter two are due to ambiguous phrasal verbs, illustrated in \ref{ex:ebol}--\ref{ex:ebol-xen}:

\vspace{-3pt}
\exg.\small\begin{coptic}a\0f\0wy\end{coptic}{~~~}\begin{coptic}~ebol\end{coptic}~\begin{coptic}jen\0ou\0cmh\end{coptic}~~\begin{coptic}m\0prop1htikon\end{coptic}\\
a-f-\={o}-\v{s}{\quad}~~ebol{~~~}xen-u-sm\={e}{~~}m-proph\={e}tikon \\
\small PST-he-cry out{\quad}~~in-a-voice{\quad~~~}of-prophetic \\
`He called out in a prophetic voice'\label{ex:ebol}

\exg. \small \begin{coptic}et\0a\0pi\0alou\end{coptic}\quad~\begin{coptic}i\end{coptic}\quad\quad\begin{coptic}ebol\end{coptic}\quad\begin{coptic}jen\0p\0hi\end{coptic}\\
    et-a-pi-alu{\quad}{\quad}{~i}{\quad}{\quad}ebol{\quad}xen-p-\={e}i \\
    \small PRC-PST-the-boy came ~~out \quad in-the-house\\
    `after the boy came out of the house'\label{ex:ebol-xen}

Like English phrasal verbs, some Coptic verbs combine with postposed adverbs to form a unique meaning -- for example in \ref{ex:ebol}, cry + out means `cry out, call out' much like in English. In the example, `out' is coincidentally followed by the preposition `in', for the phrase ``cry out in a prophetic voice''. However Coptic also has several frequent fixed combination of adverbs with following prepositions, such as \begin{coptic}ebol\end{coptic} \begin{coptic}jen\end{coptic}, lit. `out + in' but actually meaning `out of' (similar to the English fixed expression `out of' sometimes spelled as `outta'). In these cases, our guidelines annotate the second token as \texttt{fixed}, and the first token takes on the expected \texttt{case} label -- this ambiguity of adverbs next to prepositions, in which adverbs may belong together with a verb or be part of a multi-word preposition, causes the relatively frequent label confusion in Figure \ref{fig:confmat}. We note that in the interest of comparability with other languages and following UD guidelines (see \citealt{ahrenberg-2024-fitting} for discussion), the list of fixed expressions of this kind is kept small and is meant to be exhaustive, covering as of writing 25 unique combinations of lemma pairs.

\section{Conclusion}

In this paper we presented the first morphosyntactically annotated corpus of Bohairic Coptic, containing over 30K word forms from a range of texts. By adopting the same guidelines as the existing UD Sahidic Coptic Treebank, we have been able to perform some first studies of more subtle quantitative differences between the dialects, complementing the better known categorical differences between them. We also ran parsing experiments which indicated that models trained on both dialects jointly were able to boost performance on the lower resource Bohairic dialect, but not on Sahidic. 

We are hopeful that this corpus will represent a starting point for further expansion of annotated data for Bohairic Coptic in particular, and Coptic dialects in general. We are confident there is much room for both improving NLP tools for Coptic using such data, and for studying Coptic dialects individually and comparatively.

\section*{Limitations}

By its nature, this study is based on specific texts which lead to specific results. Although the attempt has been made to select somewhat diverse texts for the corpus, it is always possible that a different selection would have led to different results. In particular, the inclusion of translated texts, such as material from the Bible, is not ideal for some types of research, but as is often the case in resources for historical languages with limited attestation, this is somewhat inevitable. We are hopeful that as new data becomes available, additional studies may revisit some of our findings and either validate or relativize these results.

\section*{Acknowledgments}

This work was made possible by a grant from the National Endowment for the Humanities Preservation and Access Humanities Collections Reference and Resources Program (PW-290519-23). We thank our postdoctoral fellow, Lydia Bremer-McCollum, for digitizing some of the text used in this paper via OCR, and Wolfgang Jentner and the Data Institute for Societal Challenges at the University of Oklahoma for their server support. We would also like to acknowledge work done by other members of the Coptic Scriptorium project, the Marcion project and the Coptic Old Testament project, as well as by Hany Takla of the St. Shenouda the Archimandrite Coptic Society, which was instrumental in making the data we annotated in this paper available in digital formats. All resources developed in this project will be released under an open license in the hope that we can contribute and match their generosity.

\bibliography{acl_latex}

\appendix

\section{Dependency relations}\label{sec:appendix-labels}

We use the entire inventory of Universal Dependency relations with the exception of the \texttt{clf} relation, since Coptic has no classifiers, and no cases of an underspecified \texttt{dep} relation, for a total of 32 basic relations. In addition, we use the following four subtypes, as used in the Sahidic treebank:

\begin{itemize}
    \item \texttt{acl:relcl} - to distinguish relative clauses from adnominal infinitives and other adnominal clauses
    \item \texttt{nmod:poss} - for adnominal possessive pronouns, including both enclitic pronoun possessors and prenominal possessive pronouns
    \item \texttt{nmod:unmarked} - for adnominal, adverbially used noun phrases, not mediated by a preposition
    \item \texttt{nmod:unmarked} - for adverbially used noun phrases, not mediated by a preposition, when modifying a verbal head
\end{itemize}

We do not use the subtype \texttt{nsubj:pass} since Coptic has no unambiguous actional passive, instead using impersonal third person active syntax (``they built it'' = ``it was built''). The total distinct labels in the corpus therefore number 36.

\section{Hyperparameters}
\label{sec:appendix-hyper}

The following hyperparameters were used for DiaParser, based on the default parameters combined with the embeddings size of the MicroBERT transformer model:

\begin{itemize}
\setlength\itemsep{0.1em}
    \item BertEmbedding
    \begin{itemize}
        \item n\_layers=4
        \item n\_out=100
        \item max\_len=512
    \end{itemize}
    \item embed\_dropout: p=0.33
    \item LSTM
    \begin{itemize}
        \item dimensions: 200 x 400 x 3 layers
        \item bidirection=True
        \item dropout=0.33
    \end{itemize}
    \item MLP dropouts (arc\_d/h, rel\_d/h): 0.33
    \item criterion=CrossEntropyLoss
\end{itemize}

\end{document}